\documentclass[10pt,twocolumn,letterpaper]{article}

\usepackage{iccv}
\usepackage{times}
\usepackage{graphicx}
\usepackage{epsfig}
\usepackage{microtype}
\usepackage{amsmath}
\usepackage{amssymb}
\usepackage{subfigure}
\usepackage{booktabs} 
\usepackage{times}  
\usepackage{helvet}  
\usepackage{url}  
\usepackage{nameref}
\usepackage{amsfonts,bm,color,soul}
\usepackage{amsopn}
\usepackage{amsmath}
\usepackage{amssymb}
\usepackage{empheq}
\usepackage{epstopdf}
\usepackage{float}
\usepackage{framed}
\usepackage{adjustbox}
\usepackage{subfigure}
\usepackage{multirow}
\usepackage{dcolumn}
\usepackage{makecell}
\usepackage[para,online,flushleft]{threeparttable}
\usepackage{booktabs}
\usepackage{ulem}
\usepackage[title]{appendix}

\newcommand{\argmin}{\operatornamewithlimits{arg\,min}}
\newcommand{\argmax}{\operatornamewithlimits{arg\,max}}

\DeclareMathOperator*{\minimize}{\text{minimize}}

\DeclareMathAlphabet\mathbfcal{OMS}{cmsy}{b}{n}
\newcommand{\Def}[0]{\mathrel{\mathop:}=}
\makeatletter
\newcommand\footnoteref[1]{\protected@xdef\@thefnmark{\ref{#1}}\@footnotemark}
\makeatother

\usepackage[breaklinks=true,bookmarks=false]{hyperref}

\iccvfinalcopy 

\def\iccvPaperID{****} 

\ificcvfinal\fi

\begin{document}

\title{On the Design of Black-box Adversarial Examples by Leveraging \\ Gradient-free Optimization and Operator Splitting Method}

\author{
Pu Zhao$^{1}$, Sijia Liu$^{2}$, Pin-Yu Chen$^{2}$, Trong Nghia Hoang$^{2}$, Kaidi Xu$^{1}$, Bhavya Kailkhura$^{3}$, Xue Lin$^{1}$\\
$^1$Department of Electrical and Computer Engineering, Northeastern University\\
$^2$MIT-IBM Watson AI Lab, IBM Research\\
$^3$Syracuse University
\\
zhao.pu@husky.neu.edu, \{sijia.liu, pin-yu.chen, nghiaht\}@ibm.com,  xu.kaid@husky.neu.edu, \\ 
bkailkhu@syr.edu,  
xue.lin@northeastern.edu\\ 
}

\maketitle
\ificcvfinal\thispagestyle{empty}\fi

\begin{abstract}
Robust machine learning is currently one of the most prominent topics  which could potentially help shaping a future of advanced AI platforms that not only perform well in average cases but also in worst cases or adverse situations. Despite the long-term vision, however, existing studies on black-box adversarial attacks are still restricted to very specific settings of threat models (e.g., single distortion metric and restrictive assumption on target model's feedback to queries) and/or suffer from prohibitively high query complexity. To push for further advances in this field, we introduce a general framework based on an operator splitting method, the alternating direction method
of multipliers (ADMM)  to devise efficient, robust black-box attacks that work with various distortion metrics and feedback settings without incurring high query complexity. Due to the black-box nature of the threat model,   the proposed ADMM solution framework is integrated with zeroth-order (ZO) optimization and Bayesian optimization (BO), and thus is applicable to the gradient-free regime. This results in two new black-box adversarial attack generation methods, ZO-ADMM and BO-ADMM.
Our empirical evaluations on image classification datasets show that our proposed approaches have  much lower  function query complexities compared to state-of-the-art attack methods, but achieve very competitive attack success rates. Codes are available at \url{https://github.com/LinLabNEU/Blackbox_ADMM}.
\end{abstract}

\section{Introduction}
\label{intro}
In recent years,  deep neural networks (DNNs) have achieved  significant  breakthroughs \cite{lecun2015deep} in many machine learning (ML) tasks. 
However, despite these success stories, there have been many recent studies showing that even state-of-the-art DNNs might still be vulnerable to adversarial misclassification attacks \cite{goodfellow2015explaining,szegedy2013intriguing,xu2018structured}. The adversarial  attacks  find and add visually imperceptible noises to an originally correctly classified input and essentially cause it to be misclassified by the DNNs. This raises security concerns about the robustness of DNNs in extreme situations 
with  high reliability and dependability requirement such as face recognition, autonomous driving car and malware detection \cite{mahmood2017adversarial,evtimov2017robust,weiwei2018black}. Investigating adversarial examples has  become an increasingly prevailing topic to develop potential defensive measures in trustworthy ML \cite{lin2019defensive,wang2018defensive,xu2019topology,Wang2019HRS}. 
\textcolor{black}{It essentially lays the groundwork for building a new generation of highly robust and reliable ML models acting as the core engine of future AI technology.} 

However, most of  preliminary studies on this topic are  restricted to the white-box setting where  the adversary has complete access and knowledge of the target system (e.g., DNNs) \cite{goodfellow2015explaining,kurakin2016adversarial,carlini2017towards,chen2017ead,xu2019interpreting}. 
Despite the theoretical interest, white-box attack methods are not adapted to 
practical black-box threat models. It is often the case that internal states/configurations and operating mechanism of public ML systems are not revealed to the practitioners (e.g., Google Cloud Vision API). Accordingly, the only mode of interaction with the system is via submitting inputs and receiving the corresponding predicted outputs.

To boost the practicality of such approaches, a few recent works have introduced a new class of threat models that exploit either a surrogate of the target model \cite{nicolas2016transferability} or a
gradient-free  attack method \cite{chen2017zoo,arjun2017exploring}. 
However, adversarial attacks that exploit a surrogate of the target model tend to yield low success rate if the surrogate is inaccurate. On the other hand, while attacks that use zeroth-order gradient estimation \cite{chen2017zoo} are often more effective, they require a large number of queries to obtain an accurate estimate. Thus they are usually not economically efficient, especially in query-limited settings due to budget constraints. 

To mitigate the above limitations of the existing literature, \textcolor{black}{this paper introduces a new perspective
in the design of black-box adversarial attacks: We propose a general attack framework based on an operator splitting method, the alternating direction method of multipliers (ADMM), which integrates with both zeroth-order (ZO) optimization and Bayesian optimization (BO).}
Furthermore, unlike previous works which for ease of optimization often assume a specific distortion metric between an input and its perturbed version, our proposed framework is amenable to a broad family of distortion metrics including those previously used in the literature. 

{\noindent\bf Our Contributions:} 

\noindent $\bullet$
\textcolor{black}{We propose a general black-box adversarial  attack framework via  ADMM, including zeroth-order ADMM (ZO-ADMM)  and   ADMM with Bayesian optimization (BO-ADMM).}
We exploit a promising ZO-ADMM with random gradient estimation (RGE) \cite{liu2017zeroth} to design efficient black-box attacks that generalize the previous ZO coordinate descent based black-box attacks \cite{chen2017zoo} 
 and sidestep the notoriously intensive query complexity of attacks based on coordinate-wise random gradient estimation.
Besides, we integrate the ADMM with BO for higher query efficiency in black-box settings (Section~\ref{zoadmm}).

\noindent $\bullet$ We further generalize our formulation to accommodate various bounded $\ell_p$-norm-ball distortion metrics and their linear spans in the metric space (see Section~\ref{zstep}). Such an extension is highly non-trivial  to be incorporated into existing formulations of other black-box  attacks,
which are often heavily customized towards a specific norm-ball (e.g., $\ell_2$ or $\ell_\infty$) for distortion metrics.

\noindent $\bullet$ Our framework is also made flexible to robustly accommodate for various  threat models of the black-box attack
(Section~\ref{score-decision}), which includes both score-based and decision-based settings. The former allows the attacker to have access to a vector of assessment scores for all output candidates (soft labels). And the latter only provides the system's final decision on the most probable output (hard labels).

\noindent $\bullet$ Finally, we demonstrate the efficiency of our proposed framework on a variety of real-world image classification datasets such as MNIST, CIFAR-10 and ImageNet. The empirical results consistently show that our framework perform competitively to existing works in terms of the attack success rate while achieving a significant reduction on the query complexity (Section~\ref{experiment}).

\section{Related Works}
\label{literature}
The vulnerability of DNNs was first studied in the seminal works \cite{biggio2013evasion,szegedy2013intriguing},
which were followed by a series of white-box threat models \cite{goodfellow2015explaining,carlini2017towards,moosavi2016deepfool,zhao2018admm,pu2019fault} that assume full access of the target model's internal parameters/configurations. However, such internal knowledge of the target model is often not revealed and the adversary can only interact with it via submitting input queries and receiving feedback on potential outputs. 
Therefore, in the remaining of this section, we will summarize recent advances on black-box adversarial attacks and discuss their limitations in comparison to our proposed framework.

\subsection{Black-box Attack with Surrogate Model}
\label{surrogate}
A black-box attack using surrogate model is essentially a transfer attack \cite{nicolas2016transferability} in which the adversary trains a DNN with data labeled by the target model. The resulting DNN is then exploited as a surrogate of the target model for which we can apply any state-of-the-art white-box attacks 
without requiring full access to internal states and operating mechanisms of the target model. Such attacks however depend heavily on the quality of training a surrogate model that closely resembles the true target model \cite{yanpei2017delving}. As a result, transfer attack tends to yield low success rate in data-intensive domains (e.g., ImageNet) for which it is hard to find a qualifiled surrogate.


\subsection{Black-box Attacks with Gradient Estimation }
\label{gradient}
Another approach to explore black-box attacks is to use gradient estimation via zeroth-order optimization (ZOO) \cite{chen2017zoo}. They make queries to the model and estimate the output gradients with respect to the corresponding inputs, and then apply the state-of-the-art C\&W attack method \cite{carlini2017towards} to generate adversarial examples. However, this method is very computationally intensive as it requires a large number of queries per iteration to generate an accurate gradient estimation. Alternatively, the work \cite{nina2016simple} 
aims to estimate output gradient via greedy local search. At each iteration, the proposed technique perturbs only a small subset of input component. Such local search technique is very computationally efficient but it does not explicitly minimize the distortion between the original input and its perturbed version, the crafted noises often appear more visible. 
The work \cite{ilyas2018blackbox} investigates the   more realistic threat models by defining the query-limited setting, the partial information setting, and the label-only setting. Three attacks methods  are proposed based on the  Natural Evolutionary Strategies and Monte Carlo approximation. But it only puts limits on the $\ell_\infty$ norm instead of minimizing a certain $\ell_p$ norm. Based on \cite{ilyas2018blackbox}, the work \cite{ilyas2018prior} further investigates to utilize the prior information including the time-dependent priors (i.e., successive gradients are in fact heavily correlated) and the data-dependent priors (i.e., images tend to exhibit a spatially local similarity ) for higher query efficiency. 


\subsection{Other Black-box Attacks}
\label{others}
In addition to the aforementioned works, there are also other  black-box attacks \cite{brendel2017decision,tu2018autozoom,cheng2018queryefficient,bhavya_uni} under different practical settings, which are explored very recently. Among those, the notable boundary method
\cite{brendel2017decision}
implements a decision-based attack, which starts from a very large adversarial perturbation (thus causing an immediate misclassification) and tries to reduce the perturbation (i.e., minimize the distortion) through a random walk while remaining adversarial via staying on the boundary between the misclassified class and the true class. 
However, it suffers from high computational complexity due to a huge number of queries needed to decrease the distortion and it also has no guarantee on the convergence. 
Different from \cite{brendel2017decision}, the work \cite{cheng2018queryefficient} formulates the hard-label black-box attack as a real-valued optimization problem which is usually continuous and can be solved by the zeroth-order optimization algorithm. Similarly, \cite{bhavya_uni} addresses the problem of finding a universal (image-agnostic) perturbation in the hard-label black-box setting.

In this paper, we will instead introduce an interesting reformulation of adversarial black-box attack based on ADMM, \textcolor{black}{including ZO-ADMM \cite{liu2017zeroth} that enjoys the operator splitting advantage of ADMM and BO-ADMM that reduces the query complexity with the aid of Gaussian process.}

\section{Problem Formulation}
\label{formulation}
In this work, we   focus on adversarial attacks in the application of image classification with DNNs. In what follows, we first provide a general problem formulation for adversarial attack which is amenable to either white-box or black-box settings. Then, we will develop an efficient solution to the more interesting black-box setting where the adversary only has access to certain types of output of the DNN model (its internal structures and configurations are unknown to the adversary). Specifically, given a legitimate image $\mathbf x_0 \in \mathbb R^d$ with its correct class label $t_0$, we aim to design an optimal adversarial perturbation $\bm{\delta} \in \mathbb R^d$ so that the perturbed example $(\mathbf x_0 + \boldsymbol{\delta})$ is misclassified to target class $t \neq t_0$ by the DNN model trained on legitimate images. The adversarial perturbation $\boldsymbol \delta$ can be obtained by solving the  problem of the  generic form,
\begin{align}\label{eq: prob}
\begin{array}{ll}
    \displaystyle \minimize_{\boldsymbol \delta } & f(\mathbf x_0 + \boldsymbol \delta, t) + \gamma D(\boldsymbol \delta) 
    \\
  \text{subject to}    &  (\mathbf x_0 + \boldsymbol \delta) \in [0,1]^d, ~ \| \boldsymbol \delta \|_\infty \leq \epsilon,
\end{array}
\end{align}
where $f(\mathbf x, t)$ denotes an attack loss incurred by misclassifying $(\mathbf{x}_0 + \boldsymbol{\delta})$ to target class $t$,   $D(\boldsymbol{\delta })$ is a distortion function that controls perceptual similarity between a legitimate image and an adversarial example, 
and
 $\| \cdot \|_\infty$ signifies  the $\ell_\infty$ norm.
In problem \eqref{eq: prob}, the `hard' constraints ensure that the perturbed noise $\boldsymbol{\delta}$ at each pixel (normalized to $[0,1]$) is imperceptible up to a predefined $\epsilon$-tolerant threshold, 
and the non-negative 
parameter $\gamma$ places emphasis on the distortion. 
Furthermore, in the above problem, we mainly set $D(\boldsymbol{\delta}) = \| \boldsymbol{\delta} \|_2^2$, which is motivated by the superior performance of the outstanding C\&W $\ell_2$ adversarial attack. \textcolor{black}{We highlight that $D(\bm \delta)$ can take other forms of $\ell_p$ norms as discussed in Section \ref{zstep}.}


The problem \eqref{eq: prob} is the general form of the problem in \cite{ilyas2018blackbox,ilyas2018prior} which does not consider the $D(\bm \delta)$ term. The advantage is that we are able to minimize the $\ell_p$ distortion after the adversarial perturbation is obtained, thus keeping the perturbation imperceptible. More specifically, if $\epsilon$ is too small, we may not be able to obtain a successful adversarial example. Thus, we need to increase $\epsilon$ to achieve a successful adversarial attack. But since $\epsilon$ only limits the largest element of the perturbation, the whole perturbation over the image might be relatively large and easy to be recognized in case of large $\epsilon$.  Thus, the $D(\bm \delta)$ term in problem \eqref{eq: prob} helps to minimize the  $\ell_p$ distortion of the whole perturbation, keeping it unnoticeable. 

In the remaining of this section, we will discuss possible choices for the loss function $f(\mathbf x, t)$. Note that, without loss of generality, we only focus on  targeted attack with designated target class $t$ to mislead the DNN since the untargeted attack version can be  easily  implemented similar to the targeted attack \cite{carlini2017towards}.
We also  emphasize that in the black-box setting, the gradients of $f(\mathbf x, t)$ can not be obtained directly as it does in the white-box setting.
The form of the loss function  $f(\mathbf x, t)$ depends on the  constrained information in different black-box feedback settings. In particular, the definition of score-based (Section~\ref{score}) and decision-based (Section~\ref{decision}) attacks as well as their loss functions will be discussed in the following subsections.

\subsection{Score-based Attack}
\label{score}
In the score-based attack setting, the  adversaries are able to  make queries to DNN to obtain the  soft labels (i.e., scores or probabilities of an image  belonging to different classes), while information on gradients are not available. 
The loss function of problem \eqref{eq: prob}  in the score-based attack is:
\begin{align}\label{eq: fx}
f(\mathbf x_0 + \boldsymbol \delta, t) \ =\ \max \{ & \max_{j \neq t} \{ \text{log} P(\mathbf x_0+ \boldsymbol \delta)_j \} \nonumber
\\ & - \text{log} P(\mathbf x_0+ \boldsymbol \delta )_t, - \kappa \},
\end{align}
\textcolor{black}{which is motivated by \cite{carlini2017towards}} and yields the best known performance among white-box attacks. $P(\mathbf x)_j$ denotes the target model's prediction score or probability of the $j$-th class,  and $\kappa$ is a confidence parameter which is usually set to zero.  
Basically, this implies $f(\mathbf x_0 + \boldsymbol \delta, t) = 0$ if  $P(\mathbf x_0+ \boldsymbol \delta )_t$ is the largest among all classes, which means the perturbation $\bm \delta$ has successfully made the target model misclassified $\mathbf{x}_0 + \boldsymbol{\delta}$ to target \textcolor{black}{class} $t$. Otherwise, it will be larger than zero. 
Note that in Eqn.~\eqref{eq: fx} the log probability $\log P(\mathbf x)$ is used instead of directly using the actual probability $P(\mathbf{x})$. This is based on the observation that   the output probability distribution  
tends to have one dominating class,  making the query on the  probability/score less effective.  
The utilization of the log operator can help to reduce the effect of the  dominating class while it  preserves the probability order for all classes. 



\subsection{Decision-based Attack}
\label{decision}
Different from the score-based attack, the decision-based attack is more challenging in  that the adversaries can only make queries to get the hard-labels instead of the soft-labels. 
Let $H(\mathbf x)_i$ denote the  hard-label decision. 
$H(\mathbf x)_i = 1$ if the decision for $\bm x$ is label $i$, and $0$ otherwise. We also have $\sum_{i=1}^K H(\mathbf x)_i = 1$ for   all $K$ classes. Then  the loss function of problem \eqref{eq: prob}  in the decision-based attack  is specified as
\begin{align}\label{eq: fx_hard}
f(\mathbf x_0 + \boldsymbol \delta, t) =\max_{j \neq t}  H(\mathbf x_0+ \boldsymbol \delta)_j -H(\mathbf x_0+ \boldsymbol \delta )_t,
\end{align}
 Therefore, $f(\mathbf x_0 + \boldsymbol \delta, t) \in \{-1,1 \}$, and the attacker succeeds if $f(\mathbf x_0 + \boldsymbol \delta, t) = -1$.
The loss function (\ref{eq: fx_hard}) is \textit{nonsmooth} with \textit{discrete outputs}.  The decision-based attack is therefore more challenging because existing combinatorial optimization methods become almost ineffective or inapplicable.


\section{A General Black-box Adversarial Attack Framework}
\label{zoadmm}

This section develops a general black-box adversarial attack framework for both the score-based and decision-based attacks by leveraging  ADMM and gradient-free optimization. We will show that the proposed attack  framework yields the following benefits: a) an efficient splitting between the black-box loss function  and the adversarial distortion function,  b)  generalization to various $\ell_p$ norm involved hard/soft constraints, and \textcolor{black}{c) compatibility to different gradient-free operations.}
By introducing an auxiliary variable $\mathbf z$, problem \eqref{eq: prob}  can be rewritten in the favor of ADMM-type methods \cite{boyd2011distributed,pu2019admm},
\begin{align}\label{eq: prob_score_admm}
\begin{array}{ll}
    \displaystyle \minimize_{\boldsymbol \delta, \mathbf z } & f(\mathbf x_0 + \boldsymbol \delta, t) + \gamma D(\mathbf z) + \mathcal I(\mathbf z)\\
  \text{subject to}    &  \mathbf z = \boldsymbol{\delta},
\end{array}
\end{align}
where $\mathcal I(\mathbf z)$ is the indicator function given by,
\begin{align}
 \mathcal I(\mathbf z) = \left \{  
 \begin{array}{ll}
 0        &      (\mathbf x_0 + \mathbf z) \in [0,1]^d, ~ \| \mathbf z \|_\infty \leq \epsilon,\\
     \infty    &  \text{otherwise}.
    \end{array}
    \right.
\end{align}

The augmented Lagrangian of \textcolor{black}{the reformulated} problem \eqref{eq: prob_score_admm} is given by
\begin{align}
    \mathcal L(\mathbf z, \boldsymbol{\delta}, \mathbf u) = 
   \gamma D(\mathbf z) + \mathcal I(\mathbf z)+  f(\mathbf x_0 + \boldsymbol \delta, t) \\    \nonumber
  + \mathbf u^T(\mathbf z - \boldsymbol{\delta}) + \frac{\rho}{2} \|  \mathbf z - \boldsymbol{\delta} \|_2^2,
\end{align}
where  $\mathbf u$ is Lagrangian multiplier, and $\rho > 0$ is a given penalty   parameter. It can be further transformed as below,
\begin{align}
    \mathcal L(\mathbf z, \boldsymbol{\delta}, \mathbf u) = 
   \gamma D(\mathbf z) + \mathcal I(\mathbf z)+  f(\mathbf x_0 + \boldsymbol \delta, t) \\    \nonumber
  + \frac{\rho }{2}\left\| {\bm{z}  - \bm{\delta} + \frac{1}{\rho}\bm{u}} \right\|_2^2 - \frac{1 }{2\rho}\left\| \bm{u} \right\|_2^2.
\end{align}

The ADMM algorithm \cite{boyd2011distributed}   splits  optimization variables into \textit{two} blocks and 
adopts   the following iterative scheme,
\begin{align}
  &  \mathbf z^{k+1} \ =\  \argmin_{ \mathbf z }   \mathcal L(\mathbf z, \boldsymbol{\delta}^k, \mathbf u^k), \label{eq: z_step}\\
  &  \boldsymbol \delta^{k+1}\ =\ \argmin_{\boldsymbol{\delta}} \mathcal L( \mathbf z^{k+1}, \boldsymbol \delta,   \mathbf u^k ), \label{eq: delta_step} \\
  & \mathbf u^{k+1} \ =\ \mathbf u^k + \rho ( \mathbf z^{k+1} - \boldsymbol \delta^{k+1}),
  \label{eq: dual_update}
\end{align} 
where $k$ denotes the iteration index. In problem (\ref{eq: z_step}), we minimize $ \mathcal L(\mathbf z, \boldsymbol{\delta}, \mathbf u)$ over $\mathbf{z}$ given parameters $\boldsymbol{\delta}^k$ and $\mathbf u^k$. In problem (\ref{eq: delta_step}), we minimize $ \mathcal L(\mathbf z, \boldsymbol{\delta}, \mathbf u)$ over $\bm \delta$ given $\boldsymbol{z}^{k+1}$ from the previous step  and $\mathbf u^k$. Then, the Lagrangian multiplier $
\bm u$ is updated in Eqn.~\eqref{eq: dual_update}. 
The major advantage of this ADMM-type algorithm is that it allows us to split
the original complex problem into sub-problems, each of which can be
solved more efficiently or even analytically. In what follows, we solve \textcolor{black}{problems} (\ref{eq: z_step}) and (\ref{eq: delta_step}) respectively.

\subsection{$\mathbf z$-step}
\label{zstep}
Problem \eqref{eq: z_step} can be rewritten as
\begin{align}\label{eq: z_step_sim}
    \begin{array}{ll}
\displaystyle \minimize_{\mathbf z} &  D(\mathbf z) + \frac{\rho}{2\gamma} \| \mathbf z - \mathbf a \|_2^2     \\
        \text{subject to}  &   (\mathbf x_0 + \mathbf z) \in [0,1]^d, ~ \| \mathbf z \|_\infty \leq \epsilon,
    \end{array}
\end{align}
where $\mathbf a = \boldsymbol{\delta}^k - (1/\rho) \mathbf u^k $. We set $D(\boldsymbol{z}) = \| \boldsymbol{z} \|_2^2$ \cite{carlini2017towards}.
Problem \eqref{eq: z_step_sim} can be decomposed elementwise as below,  
\begin{align}\label{eq: z_step_sim_ele}
\begin{array}{ll}
    \displaystyle \minimize_{z_i}  &  \left (z_i - \frac{\rho}{2\gamma+\rho} a_i \right )^2 \\
    \text{subject to}    &  ( \left [ \mathbf x_0  \right ]_i +  z_i ) \in [0,1],  ~ |z_i| \leq  \epsilon,
\end{array}
\end{align}
where $[\mathbf x]_i$ (or $x_i$) denotes the $i$-th element of $\mathbf x$.
The solution to problem \eqref{eq: z_step_sim_ele} is then given by 
\begin{align}
\small
    [\mathbf z^{k+1}]_i = \left \{
    \begin{array}{ll}
       \min \{ 1- \left [\mathbf x_0 \right ]_i, \epsilon \}  &  
       \frac{\rho}{2\gamma+\rho} a_i >  \min \{ 1- \left [ \mathbf x_0  \right ]_i, \epsilon \}  \\
     \max \{ -  \left [ \mathbf x_0  \right ]_i, -\epsilon \}   &  
     \frac{\rho}{2\gamma+\rho} a_i <  \max \{ -  \left [ \mathbf x_0  \right ]_i, -\epsilon \}   \\
        \frac{\rho}{2\gamma+\rho} a_i & \text{otherwise}.
    \end{array}
    \right.
\end{align}

\textbf{Generalization to various $\ell_p$ norms.}
In problem \eqref{eq: z_step_sim}, in addition to the worst-case perturbation constraint $\|\boldsymbol{z} \|_\infty \leq \epsilon$, it is a common practice to set $D(\boldsymbol{z}) = \| \boldsymbol{z} \|_2^2$ to measure the similarity between the legitimate image and the adversarial example. If $D(\boldsymbol{z})$  takes other $\ell_p$ norms such as $\| \boldsymbol{z} \|_0$, $\| \boldsymbol{z} \|_1$, 
or even $\ell_p$ norm combinations like  $ \| \boldsymbol{z} \|_1 + \frac{\beta}{2} \| \boldsymbol{z} \|_2^2$ for $\beta \geq 0$, we are still able to obtain the solutions with minor modifications in the $z$-step. 
This ability is highly non-trivial for other black-box  attacks, which  are  often  heavily  customized to minimize a specific $\ell_p$ norm for distortion measure.
Due to space limitation, we explicitly show the $\bm z$-step solutions of $D(\boldsymbol{z}) = \| \boldsymbol{z} \|_0$, $D(\boldsymbol{z}) = \| \boldsymbol{z} \|_1$ and $D(\boldsymbol{z}) = \| \boldsymbol{z} \|_1 + \frac{\beta}{2} \| \boldsymbol{z} \|_2^2$ derived with proximal operators \cite{parikh2014proximal} in the supplementary material. 





\subsection{$\boldsymbol{\delta}$-step}
\label{delta}
Problem \eqref{eq: delta_step} can be written as
\begin{align}\label{eq: delta_step_ZO}
    \begin{array}{cc}
\displaystyle \minimize_{\boldsymbol{\delta}}  & f(\mathbf x_0 + \boldsymbol{\delta}, t) + \frac{\rho}{2} \| \boldsymbol{\delta} - \mathbf b \|_2^2  ,
    \end{array}
\end{align}
where $\bm b = \bm z^{k+1} + (1/\rho) \bm u^k$. In the \textcolor{black}{white-box setting}, since the gradients of $f(\mathbf x_0 + \boldsymbol{\delta}, t)$ are \textcolor{black}{directly accessible}, gradient descent method like stochastic gradient descent (SGD) or Adam can be applied straight-forwardly. 
However, in black-box  settings, the gradients of $f(\mathbf x_0 + \boldsymbol{\delta}, t)$ are unavailable. \textcolor{black}{  Thus, to overcome this difficulty, we adopt two derivative-free methods: the random gradient estimation (RGE) method \cite{duchi2015optimal} and the Bayesian optimization \cite{bogunovic2016truncated} corresponding to ZO-ADMM and BO-ADMM, respectively.}

\subsubsection{Random gradient estimation}
In the black-box setting, the gradient of $f(\mathbf x_0 + \boldsymbol{\delta}, t)$ is estimated through random gradient estimation (RGE),
 \begin{equation}\label{eq: grad_rand_ave}
\hat \nabla f(\boldsymbol{\delta}) = (d/(\nu Q)) \sum_{j=1}^Q 
 \Big [  f ( \boldsymbol{\delta} + \nu \mathbf u_{j} ) - f ( \boldsymbol{\delta} ) \Big ]\mathbf u_{j}, 
\end{equation}
where  $d$ is the number of optimization variables, $\nu> 0$ is a smoothing parameter, $\{ \mathbf u_j \}$ denote
 independent and identically distributed (i.i.d.) random direction vectors drawn from a uniform distribution over a unit sphere, and   $Q$ is the number of random direction vectors. 
It has been shown in \cite{liu2017zeroth} that a large $Q$ reduces the gradient estimation error and improves the convergence of ZO-ADMM. However, we find that a moderate size of $Q$ is sufficient to provide a good trade-off between estimation error and query complexity, e.g., $Q=20$ in our experiments. We also highlight that the RGE in \eqref{eq: grad_rand_ave} only requires $O(Q)$ query complexity instead of $O(dQ)$ caused by coordinate-wise gradient estimation used in \cite{chen2017zoo}. \textcolor{black}{Note that the  natural evolutionary strategy (NES) \cite{ilyas2018blackbox} uses  a central difference based gradient estimator requiring $2Q$ queries. By contrast, RGE  uses a forward difference based random gradient estimator, yielding $Q+1$ query counts, leading to higher query efficiency.}

With the aid of RGE, the solution to problem (\ref{eq: delta_step_ZO}) can now be obtained via stochastic gradient descent-like methods.
However, it suffers from extremely high iteration and function query complexity due to the non-linearity of $f$ as well as the iterative nature of ADMM.
To sidestep this computational bottleneck, we propose the use of the linearized ADMM algorithm \cite{suzuki2013dual} in ZO-ADMM with  RGE, and thus it enjoys dual advantages of gradient-free operation and linearization of the loss function.
By linearization, the loss function $ f(\mathbf x_0 + \boldsymbol{\delta}, t) $ in problem (\ref{eq: delta_step_ZO}) is replaced with  its  first-order  Taylor  expansion  plus  a  regularization  term  (known  as  Bregman divergence), that is, $\hat \nabla f(\boldsymbol \delta^k + \mathbf x_0, t))^T ( \boldsymbol \delta - \boldsymbol \delta^k) + \frac{1}{2} \| \bm \delta  - \bm \delta^k \|_{\mathbf G}^2$, where $\mathbf G$ is a pre-defined positive definite matrix, and $ \| \mathbf x \|_{\mathbf G}^2 =  \mathbf x^T \mathbf G \mathbf x $. We choose $\mathbf G = \eta_k  \mathbf I$  where $1/\eta_k >0 $ is a decaying parameter, e.g.,   $\eta_k = \alpha \sqrt{k}$ for a given constant $\alpha > 0$. The Bregman divergence term is used to stabilize the convergence of $\bm \delta$.

Combining linearization and RGE, problem \eqref{eq: delta_step_ZO} now takes the following form:
\begin{align}
    \begin{array}{ll}
\displaystyle \minimize_{\boldsymbol{\delta}}         &  \displaystyle (\hat \nabla f(\boldsymbol \delta^k + \mathbf x_0, t))^T ( \boldsymbol \delta - \boldsymbol \delta^k) \\ 
& + \frac{\eta_k}{2} \| \boldsymbol \delta - \boldsymbol \delta^k \|_2^2 + \frac{\rho}{2} \| \boldsymbol \delta - \mathbf  b \|_2^2 ,
    \end{array}
    \label{eq: delta_step_ZO_lin}
\end{align}
which yields a quadratic \textcolor{black}{programming problem} with the following closed-form solution:
\begin{equation}\label{eq: delta_sol_score}
\small
  \boldsymbol{\delta}^{k+1} = \left (1/ \left ( \eta_k +   \rho \right ) \right)
    \left ( \eta_k \boldsymbol{\delta}^k +\rho \mathbf b  - \hat \nabla f(\boldsymbol{\delta}^k + \mathbf x_0,t)
    \right ).
\end{equation}

Note that \textcolor{black}{Eqn. (\ref{eq: delta_sol_score}) can be calculated}
with only one step of gradient estimation, which is a significant improvement on query efficiency compared with solving problem (\ref{eq: delta_step_ZO}) using gradient descent method with thousands of random estimations.  
The convergence of the linearized ADMM for non-convex problems is proved in \cite{liu2017linearized}. 

\subsubsection{Bayesian Optimization}

In addition to RGE, BO is an alternative approach to solve problem \eqref{eq: delta_step_ZO} \cite{fnu2017query,ariafar2017admm,ariafar2019admmbo}.
We model 
$l(\boldsymbol{\delta})\Def f(\bm x_0 + \boldsymbol{\delta},t) + \frac{\rho}{2}  \| \boldsymbol{\delta} - \mathbf b \|_2^2$
as a Gaussian process with a \textit{prior} distribution $l(\cdot) \sim \mathcal N(\mu_0, \kappa(\cdot, \cdot))$, where $\mu_0 = 0$ in practice and $\kappa(\cdot, \cdot)$ is a positive definite kernel \cite{shahriari2016taking}. 
Consider a finite collection of \textit{noisy observations} $\mathcal D_k = \{ y_1, \ldots, y_k \}$, where $y_i   \sim \mathcal N(l(\boldsymbol{\delta}^i), \sigma_n^2)$, and $\sigma_n^2$ is the noise variance.
The \textit{posterior probability} of a new function 
$l(\boldsymbol{\delta})$
evaluation given $\mathcal D_k$  is a Gaussian
distribution with mean $\mu $ and variance  $\sigma $, that is $l(\boldsymbol{\delta}) | \mathcal D_k \sim \mathcal N (\mu , \sigma^2)$, where
\begin{align}\label{eq: post_f}
    \mu & = \boldsymbol \kappa^T [\mathbf K + \sigma_n^2 \mathbf I]^{-1} \mathbf y, \\
    \quad \sigma^2 & = \kappa(\boldsymbol{\delta} , \boldsymbol{\delta} ) - \boldsymbol \kappa^T [\mathbf K + \sigma_n^2 \mathbf I]^{-1}\boldsymbol \kappa,
\end{align}
$K_{ij} = \kappa(\boldsymbol{\delta}^i, \boldsymbol{\delta}^j)$, $\boldsymbol \kappa$ is a vector of covariance terms between $\{ \boldsymbol{\delta}^i \}_{i=1}^k$ and $\boldsymbol{\delta}$, namely, $\kappa_i = \kappa(\boldsymbol{\delta}^i, \boldsymbol{\delta})$. 

We choose the kernel function $\kappa(\cdot,\cdot)$  as the ARD Mat\'ern $5/2$ kernel \cite{snoek2012practical,shahriari2016taking},
\begin{align}\label{eq: kernel_paras}
    \kappa(\mathbf x, \mathbf y) &= \theta_0^2 \mathrm{exp}(-\sqrt{5} r) (1+ \sqrt{5}r + \frac{5}{3} r^2),\\
    \quad r^2 &= \sum_{i=1}^d (x_i - y_i)^2/\theta_i^2,
\end{align}
where $  \{ \theta_i \}_{i=0}^d$ are hyperparameters. Note that $\kappa(\boldsymbol{\delta} , \boldsymbol{\delta} ) = \theta^2_0$.

To determine the hyper-parameters $\boldsymbol{\theta} = \{  \{ \theta_i \}_{i=0}^d,  \sigma_n^2\}$,
we minimize the negative log marginal likelihood  $\log p(\mathcal D_k | \boldsymbol{\theta})$ \cite{shahriari2016taking},
\begin{align}\label{eq: learn_hyper}
\displaystyle \minimize_{\boldsymbol{\theta}} \  \  \displaystyle {L}(\boldsymbol{\theta})  &\triangleq \frac{1}{2}\log|\mathbf{K} + \sigma_n^2\mathbf{I}| \\ \nonumber &+  \frac{1}{2}\mathbf{y}^\top\left(\mathbf{K} + \sigma_n^2\mathbf{I}\right)^{-1}\mathbf{y}  ,
\end{align}
where $\mathbf{y} = [y_1\ y_2\ \ldots\ y_k]^\top$. This can be achieved by a standard gradient descent routine $\boldsymbol{\theta} \leftarrow \boldsymbol{\theta}  {-} \eta \partial{L}/\partial\boldsymbol{\theta}$ with a sufficiently small learning rate $\eta$.

 In the setting of BO, the solution to problem \eqref{eq: delta_step_ZO} is often acquired by maximizing the \textit{expected improvement} (EI). The EI acquisition function is defined as \cite{shahriari2016taking}
\begin{align}\label{eq: EI}
    \boldsymbol{\delta}^{k+1} &= \argmax \mathrm{EI}(\boldsymbol{\delta}) \nonumber \\  &= \argmax \mathbb E_{l(\boldsymbol{\delta})| \mathcal D_k} \left [ \left ( l^+ - l(\boldsymbol{\delta}) \right ) \mathcal I (l(\boldsymbol{\delta}) \leq l^+) \right ],
    \nonumber \\ &= \argmax_{ \boldsymbol{\delta} } ~ (l^+ - \mu) \Phi \left ( \frac{l^+ - \mu}{\sigma} \right ) + \sigma\phi\left (
    \frac{l^+ - \mu}{\sigma}
    \right ),
\end{align}
where $l^+$ denotes the best observed value, and $\mathcal I (l(\boldsymbol{\delta}) \leq l^+) = 1$ if $l(\boldsymbol{\delta}) \leq l^+$, and $0$ otherwise. $\Phi$ and $\phi$ denote the CDF and PDF of the standard normal distribution, respectively.  \textcolor{black}{We refer readers to the supplementary material for the detailed derivation of Eq. \eqref{eq: EI}.} 
We obtain $\bm \delta_{k+1}$ through the projected  gradient descent method, 
\begin{align}
 &   \hat{\boldsymbol{\delta}}^{(k+1)} = \boldsymbol{\delta}^{(k)} + \eta \nabla_{\boldsymbol{\delta} = \boldsymbol{\delta}^{(k)}} \mathrm{EI} (\boldsymbol{\delta}),  \\
 & \boldsymbol{\delta}^{(k+1)} = \mathrm{Proj}_{ \text{$(\mathbf x_0 + \boldsymbol{\delta}) \in [0,1]^d$,  $\|\boldsymbol{\delta} \|_\infty \leq \epsilon$} } \left (   \hat{\boldsymbol{\delta}}^{(k+1)} \right ). 
 \label{eq: proj_theta}
\end{align}
The projection 
is introduced to ensure the feasibility of the next query point in BO.

\section{Customized Score-based and Decision-based Black-box Attacks}
\label{score-decision}
For the score-based black-box attack,  problem \eqref{eq: prob} with loss function \eqref{eq: fx} can be naturally solved through the general ADMM  framework. 

In the decision-based black-box attack, the form of the loss function (\ref{eq: fx_hard}) is non-smooth with discrete outputs. To overcome the discontinuity in Eqn. \eqref{eq: fx_hard}, a smoothing version of \eqref{eq: fx_hard}, denoted by $f_\mu$ with smoothing parameter $\mu > 0$ \cite{gao2014information,nesterov2015random},
 is taken into consideration,
\begin{align}\label{eq: fx_hard_smooth}
    f_\mu(\mathbf x_0 + \boldsymbol \delta, t) \ =\  \mathbb E_{\mathbf u \in U_{\mathrm{b}}}\left[f(\mathbf x_0 + \boldsymbol \delta + \mu \mathbf u, t)\right],
\end{align}
where $U_{\mathrm{b}}$ is a uniform distribution within the unit Euclidean ball, or $\mathbf u$ can follow a standard Gaussian distribution \cite{ilyas2018blackbox}.
\textcolor{black}{The rational behind the smoothing technique in  \eqref{eq: fx_hard_smooth} is that
 the convolution of two functions, i.e.,
 $\int_{\mathbf u} f(\mathbf x_0 + \boldsymbol{\delta} + \mu \mathbf u, t) p(\mathbf u) d \mathbf u$,
 is at least as smooth as the smoothest
of the two original functions \cite{duchi2012randomized}. Therefore, when 
$p$ is the density of a random variable with respect to Lebesgue measure, the loss function \eqref{eq: fx_hard_smooth} is then smooth.}
In practice, we consider  
an empirical \textcolor{black}{Monte Carlo} approximation of \eqref{eq: fx_hard_smooth} 
\begin{align}\label{eq: fx_hard_smooth_approx}
    f_\mu(\mathbf x_0 + \boldsymbol \delta, t)\ \approx\ \frac{1}{N} \sum_{i=1}^N f(\mathbf x_0 + \boldsymbol \delta + \mu \mathbf u_i, t),
\end{align}
where $\{ \mathbf u_i \}$ are $N$ i.i.d. samples drawn from $U_{\mathrm{b}}$. With the smoothed loss function as expressed in Eqn. (\ref{eq: fx_hard_smooth_approx}), problem (\ref{eq: prob}) can be solved by the proposed general  ADMM framework. \textcolor{black}{
 To initialize ADMM, we initialize the perturbation $\boldsymbol{\delta}$ so that the initial perturbed image belongs to the target class, yielding a benefit of reducing query complexity compared to the initialization with an arbitrary image \cite{brendel2017decision}.}

\section{Performance Evaluation}
\label{experiment}
\begin{table*} [htb]
 \centering
  \caption{Performance evaluation of \textcolor{black}{adversarial} attacks on MNIST and CIFAR-10.}
  \label{table_score_MNIST_CIFAR}
  \scalebox{0.7}[0.7]{
   \begin{threeparttable}
\begin{tabular}{c|c|c|c|c|c|c|c|c}
    \hline
\toprule[1pt]
Data set & \multicolumn{2}{c|}{Attack method}  & 
\makecell{ASR} &  \makecell{ $\ell_1$ \\ distortion}&  \makecell{ $\ell_2$ \\ distortion}&  \makecell{ $\ell_\infty$ \\ distortion} & \makecell{\textcolor{black}{Query count on} \\ \textcolor{black}{initial success}} & \makecell{Reduction ratio on \\ query count} \\ 
\midrule[1pt]
\multirow {7}{*}{MNIST}  & \multirow {2}{*}{\makecell{ white\\-box}}  & C\&W white-box attack \cite{carlini2017towards} & 100\% & 22.14 & 1.962& 0.5194 & - & -  \\
&  & \makecell{Transfer attack (via C\&W)} \cite{nicolas2016transferability} & 30.6\% & 65.2 & 4.545 & 0.803 & - & - \\ 
\cline{2-9} 
& \multirow {3}{*}{\makecell{ score\\-based}}  & ZOO attack \cite{chen2017zoo} & 98.8\% & 26.78 &1.977& 0.522 &  12,161 & 0.0 \% \\ 
&  & score-based ZO-ADMM  attack   & 98.3\% & 26.23 & 1.975& 0.513 & 493.6 & 95.9\% \\ 
&   &  score-based BO-ADMM attack  & 87\% & 93.6 & 7.7 & 0.71 & 52.1 & 99.6\% \\ 
\cline{2-9} 
& \multirow {2}{*}{\makecell{ decision\\-based}}  &  boundary attack \cite{brendel2017decision} & 99\%  & 32.9  & 2.21 & 0.563 & 25,328\tnote{a}  & 0\% \\
&  & decision-based ZO-ADMM attack   & 100 \% & 30.48 &2.166 &  0.548 & 7,603 \tnote{a} & 62\%  \\ 
\midrule[1pt]
\multirow {7}{*}{CIFAR-10} & \multirow {2}{*}{\makecell{ white\\-box}}  & C\&W white-box attack \cite{carlini2017towards} & 100 \% & 11.7  & 0.332 & 0.0349 & - & - \\ 
&  & \makecell{Transfer attack (via C\&W)} \cite{nicolas2016transferability} & 8.5\% & 103.6 & 3.845 & 0.421 & - & - \\ 
\cline{2-9}  
& \multirow {3}{*}{ \makecell{ score\\-based}}   & ZOO attack \cite{chen2017zoo}  & 97.6 \% & 15.2 &0.361& 0.0405 & 9982  & 0.0 \% \\ 
& &  score-based ZO-ADMM attack  &98.7 \% & 13.1 & 0.417& 0.0392 & 421 & 95.7\%  \\ 
& &  score-based BO-ADMM attack & 84.1\% & 148 & 5.29 & 0.62 & 46.3 & 99.6\% \\
\cline{2-9} 
& \multirow {2}{*}{\makecell{  decision \\ -based}}  &  boundary attack \cite{brendel2017decision}  & 100\%  & 19.4  & 0.421 & 0.045 & 16,720 \tnote{a} & 0\%  \\ 
&  &  decision-based ZO-ADMM attack  & 100\% & 17.25 & 0.415 & 0.0413  & 6,213 \tnote{a} & 63\% \\ 
\bottomrule[1pt]
  \end{tabular}
  \begin{tablenotes}
\item[a] As the decision-based attacks start from images in the target class, it achieves initial success immediately. Therefore,  the query count on the initial success of the  decision-based  attack actually means the query number when it achieves the reported $\ell_2$ distortion.
\end{tablenotes}
\end{threeparttable}}  
\end{table*}

In this section, the experimental results of the score-based and decision-based black-box attacks are demonstrated. 
We compare the proposed  ADMM-based framework with 
various attack methods on three image classification datasets, MNIST \cite{Lecun1998gradient}, CIFAR-10 \cite{Krizhevsky2009learning} and ImageNet \cite{deng2009imagenet}. The results of state-of-the-art white-box attack (i.e., C\&W attack) are also provided for reference.

We train two networks for MNIST and CIFAR-10 datasets, respectively, which can achieve 99.5\% accuracy on MNIST and 80\% accuracy on CIFAR-10. 
The model architecture 
has four convolutional layers, two max pooling layers, two fully connected layers and a softmax layer. 
For ImageNet, we utilize a pre-trained Inception v3 network \cite{Szegedy2016RethinkingTI} instead of training our own model, which can achieve 96\% top-5 accuracy.  
All experiments are conducted on machines with 
NVIDIA GTX 1080 TI GPUs.

\subsection{Evaluation on MNIST and CIFAR-10}

In the evaluation on MNIST and CIFAR-10, 200 correctly classified images are selected from MNIST and CIFAR-10 test datasets, respectively. For each image, the target labels are set to the other 9 classes and a total of 1800 attacks are performed for each attack method.

The implementations of C\&W \textcolor{black}{(white-box)} attack \cite{carlini2017towards} and ZOO \textcolor{black}{(black-box)} attack \cite{chen2017zoo}
are based on the GitHub code released by the authors\footnote{\label{note1}
Codes are available at \url{https://github.com/LinLabNEU/Blackbox_ADMM}.
}. 
For ZOO attack, we use ZOO-Adam with default Adam parameters. 
For the transfer attack \cite{nicolas2016transferability}, we apply C\&W attack  to the surrogate model 
\textcolor{black}{with} $\kappa = 20$ to improve the attack transferability \textcolor{black}{and}  2,000 iterations in each binary search step. 
In the proposed ZO-ADMM attack\footnoteref{note1}, the \textcolor{black}{sampling} number 
in random gradient estimation \textcolor{black}{as defined in Eqn. (\ref{eq: grad_rand_ave})}, $Q$, is set to 20 and the \textcolor{black}{sampling}
number for the decision-based smoothed loss function (\ref{eq: fx_hard_smooth_approx}), $N$, is set to 10. 
\textcolor{black}{ We set $\rho=10$ and $\gamma=1$ for MNIST, $\rho=2000$ and $\gamma=10$ for CIFAR-10, and $\rho=1000$ and $\gamma=1$ for ImageNet. $\epsilon$ is set to 1  for MNIST and CIFAR-10  \footnote{\textcolor{black}{This setting is intended to make a fair comparison to the pure $\ell_2$-norm attack framework ZOO. }}  and 0.05 for ImageNet. In Eq. \eqref{eq: grad_rand_ave}, we set $\nu=0.5$   for three datasets.
The parameter $\mu$ in Eq. \eqref{eq: fx_hard_smooth_approx} is set to $1$ for MNIST, $0.1$ for CIFAR-10, and $0.01$ for ImageNet. }

\begin{figure}[htb]
 \vspace*{-0.15in}
\centering
\subfigure[\textcolor{black}{An adversarial example evolution for MNIST starting from an image in the target class.}]{
\begin{minipage}[b]{0.4\textwidth}
\includegraphics[width=0.85\textwidth]{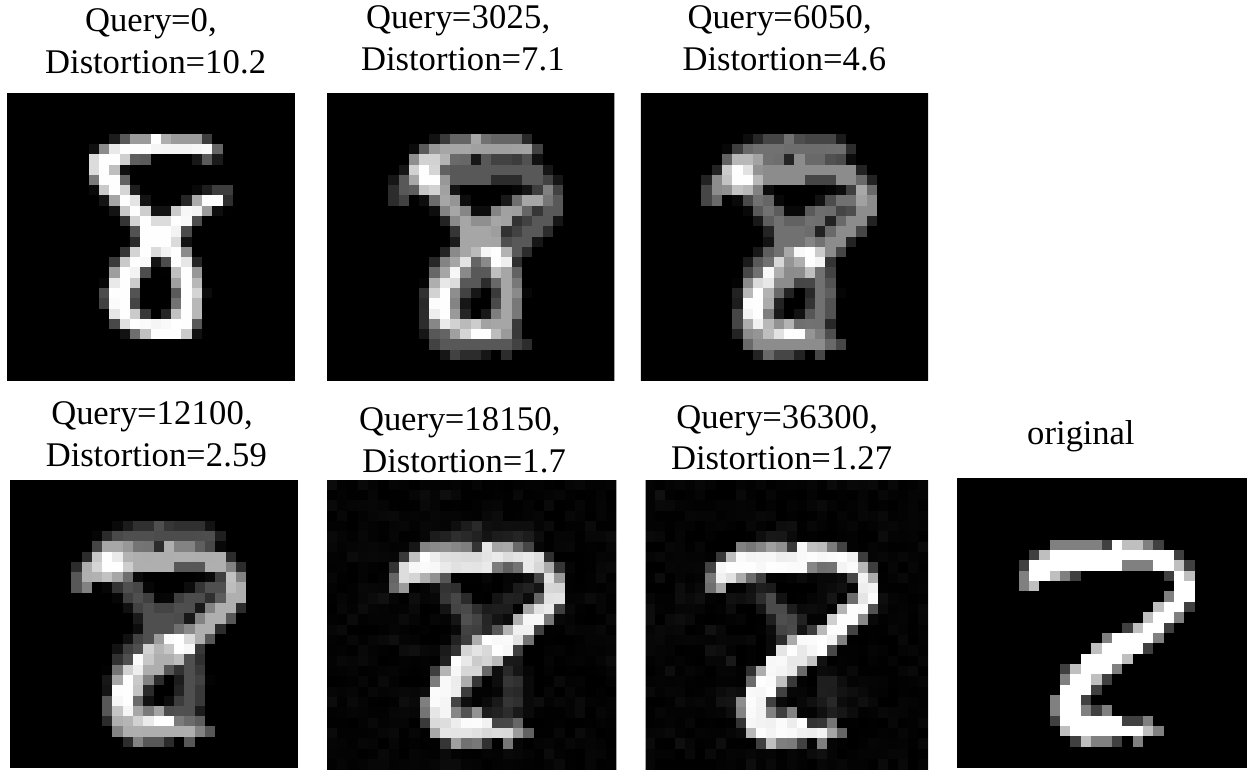}
\end{minipage}
}
\subfigure[\textcolor{black}{An adversarial example evolution for CIFAR-10 starting from an image in the target class.}]{
\begin{minipage}[b]{0.4\textwidth}
\includegraphics[width=0.85\textwidth]{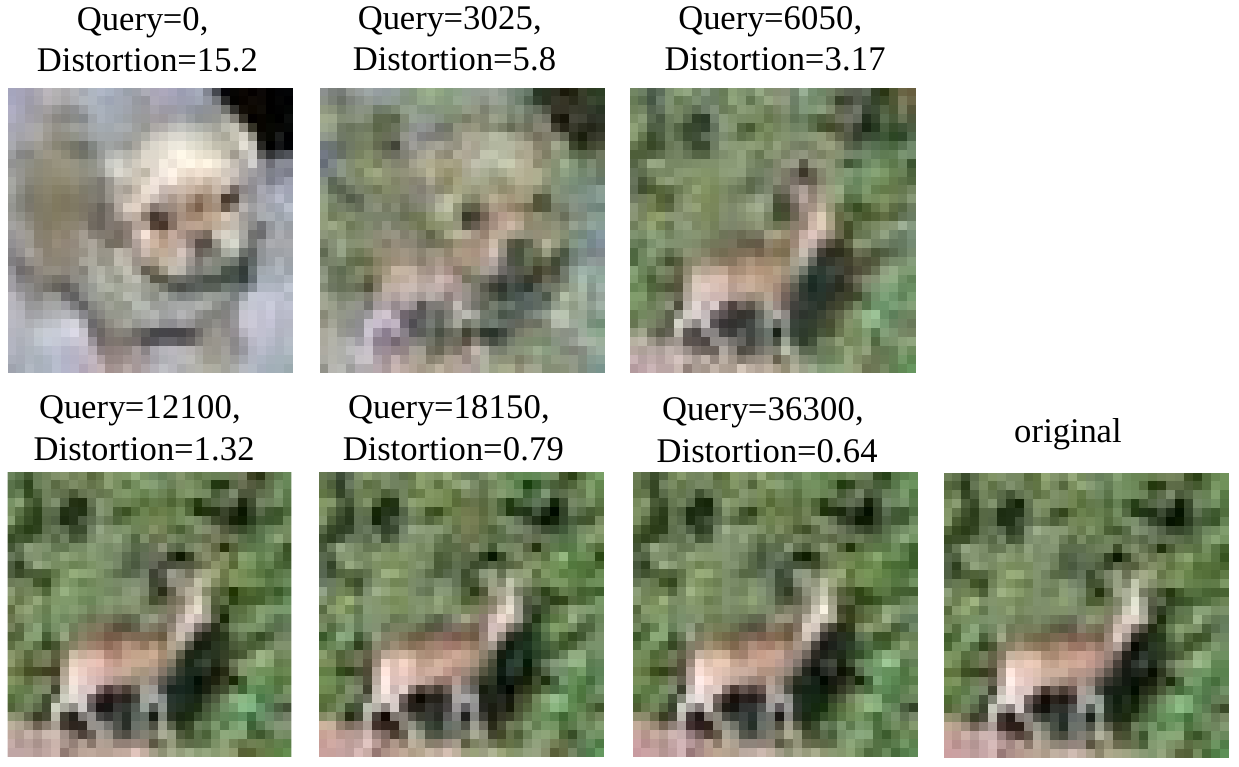}
\end{minipage}
}
 \caption{\textcolor{black}{Adversarial examples generated by the proposed decision-based ZO-ADMM  attack} on MNIST and CIFAR-10.\vspace{-4mm} } \label{envo_mnist_cifar} 
\end{figure}

The experimental results are shown in Table \ref{table_score_MNIST_CIFAR}.
\textcolor{black}{Besides the attack success rate (ASR) and the $\ell_p$ norms, we report the query number required to achieve the first successful attack, which characterizes how fast the generated adversarial perturbation can mislead DNNs.
}
We observe that the transfer attack suffers from low ASR and large $\ell_2$ distortion.
Both the ZOO attack and the proposed ZO-ADMM attack with RGE  can achieve high ASR and competitive $\ell_2$ distortion  close to the C\&W white-box  attack. 
Compared with the ZOO attack, the   score-based  ZO-ADMM attack   requires  fewer queries to obtain the first successful adversarial example. The query count  in ZO-ADMM attack with RGE is reduced by 95.9\% and 95.7\% on MNIST and CIFAR-10, respectively. 
The reduction of query number is achieved by Eq. \eqref{eq: delta_sol_score} in ZO-ADMM, which only requires one step of gradient estimation to solve the approximation problem instead of thousands of steps to solve the original problem.   
We also observe that the score-based BO-ADMM attack   can achieve smaller query number compared with the RGE method, but it causes much larger $\ell_p$ distortion. The reason is that BO-ADMM does not have very precise control for the perturbation. So it requires larger perturbation to mislead the DNN model. 
\textcolor{black}{ Although BO-ADMM may have its limitations, we find that combining the advantage of BO- and ZO-ADMM can lead to more query-efficient attacks. Please refer to the BO-ZO-ADMM section in the appendix.}

\begin{table*} [tb]
 \centering
  \caption{Performance evaluation of adversarial attacks on ImageNet.}
  \label{table_ADMM_ImageNet}
  \scalebox{0.7}[0.7]{
   \begin{threeparttable}
\begin{tabular}{c|c|c|c|c|c|c|c}
    \hline
\toprule[1pt]
\multicolumn{2}{c|}{ } &  \multicolumn{3}{c|}{Untargeted attack}  &  \multicolumn{3}{c}{Targeted attack}\\
\hline
\multicolumn{2}{c|}{ Attack method} & 
\makecell{ASR}  & \makecell{Query count on\\   initial success} & \makecell{Reduction ratio }  &
\makecell{ASR} & \makecell{Query count on \\ initial success} & \makecell{Reduction ratio } \\
\midrule[1pt]
 \multirow {4}{*}{ score-based } & C\&W white-box attack \cite{carlini2017towards}  & 100\%  & - & - & 99\% & - & - \\
& ZOO attack \cite{chen2017zoo} & 90\%  & 15631 & 0.0\% & 78\% & $2.11\times10^6$ & 0.0\%  \\
&Query-limited attack \cite{ilyas2018blackbox} & 100\% & 4785 & 69.4\% & 98\% & 34128 & 98.4\% \\ 
&  Bandits$_\text{TD}$  \  attack \cite{ilyas2018prior} & 94\% & 1259 & 92\%  & - \tnote{a} & - & -\\ 
&  score-based ZO-ADMM attack & 98\%  & 891 & 94.3\% & 97\% &  16058 & 99.2\%  \\
 \midrule[1pt]
 \multirow {2}{*}{ decision-based } & Label only \cite{ilyas2018blackbox} &  - \tnote{b}  & - &- & 92 \% & $1.89\times10^6$ \tnote{c} & 10.4\%  \\
 &  decision-based ZO-ADMM attack & 100\%  & 11742 \tnote{c} & 24.9\%  & 94\% & $1.52\times10^6$ \tnote{c} & 28\%   \\
\bottomrule[1pt]
  \end{tabular}
    \begin{tablenotes}
    \item[a] It mainly explores untargeted attack.
    \item[b] The label only attack mainly explores targeted attack.
\item[c] 
The query count on initial success for the decision-based attack means the query number when it achieves the same $\ell_2$ distortion with the ZOO attack on its initial success.
\end{tablenotes}
\end{threeparttable}} \vspace{-15pt}
\end{table*}

We notice that the  decision-based ZO-ADMM attack   achieves an $\ell_2$ distortion slightly larger than the score-based attack with more queries as shown in Table \ref{table_score_MNIST_CIFAR}. This is not surprising, since only the hard label outputs are available in the decision-based attack, which is  more difficult to be optimized than the score-based attack. Although the $\ell_2$ distortion is a bit larger, the perturbations are still visually indistinguishable.
We compare  the  decision-based ZO-ADMM attack  with the boundary attack \cite{brendel2017decision}. As demonstrated in Table \ref{table_score_MNIST_CIFAR}, the queries of the  decision-based ZO-ADMM attack is about 60\% less than that of the boundary attack to achieve the same level $\ell_2$ distortion. 
 We show the evolution of several adversarial examples in the decision-based attack versus the query number in Fig. \ref{envo_mnist_cifar}. The decision-based attack starts from an image in the target class. 
Then it tries to decrease the $\ell_2$ norm 
while keeping the classified label unchanged. After about 20,000 queries, the example is close to the original image with a satisfied $\ell_2$ distortion.

\subsection{Evaluation on ImageNet}

We perform targeted and untargeted attacks in the score-based and decision-based settings on ImageNet. 100 correctly classified images are randomly selected. For each image in targeted attack, 9 random labels out of 1000 classes are selected to serve as the targets. We do not perform the transfer attack since it does not scale well to ImageNet due to training of the surrogate model.
Instead, we provide the results of  new baselines on ImageNet, including the query-limited attack as well as the label-only  attack proposed in \cite{ilyas2018blackbox}, and the bandit optimization based attack with time and data-dependent priors (named as Bandits$_\text{TD}$) \cite{ilyas2018prior}.
The query-limited  and Bandits$_\text{TD}$ attacks are  score-based attacks. The label-only attack is a decision-based attack.

The experimental results are summarized in Table \ref{table_ADMM_ImageNet}. 
For   score-based attacks, we can observe that the  score-based ZO-ADMM attack can achieve a high ASR with fewer queries than the other attacks.
It reduces the query number on initial success by 94.3\% and 99.2\% for untargeted and targeted attacks, respectively, compared with the ZOO attack.  
For decision-based attacks, the ZO-ADMM attack can obtain a high ASR with fewer queries compared with the label-only attack or even the ZOO attack using score-based information. Some adversarial examples generated by the ZO-ADMM attack are demonstrated in the supplementary material.  \textcolor{black}{ More experimental results including the comparison with AutoZoom \cite{tu2018autozoom} and the boundary method \cite{brendel2017decision} method are demonstrated in the Appendix.}


\subsection{Convergence of the ZO-ADMM Attack} \label{convgence_analysis}

\begin{figure}[tb]
\centerline{
\begin{tabular}{c}
\includegraphics[width=.3\textwidth]{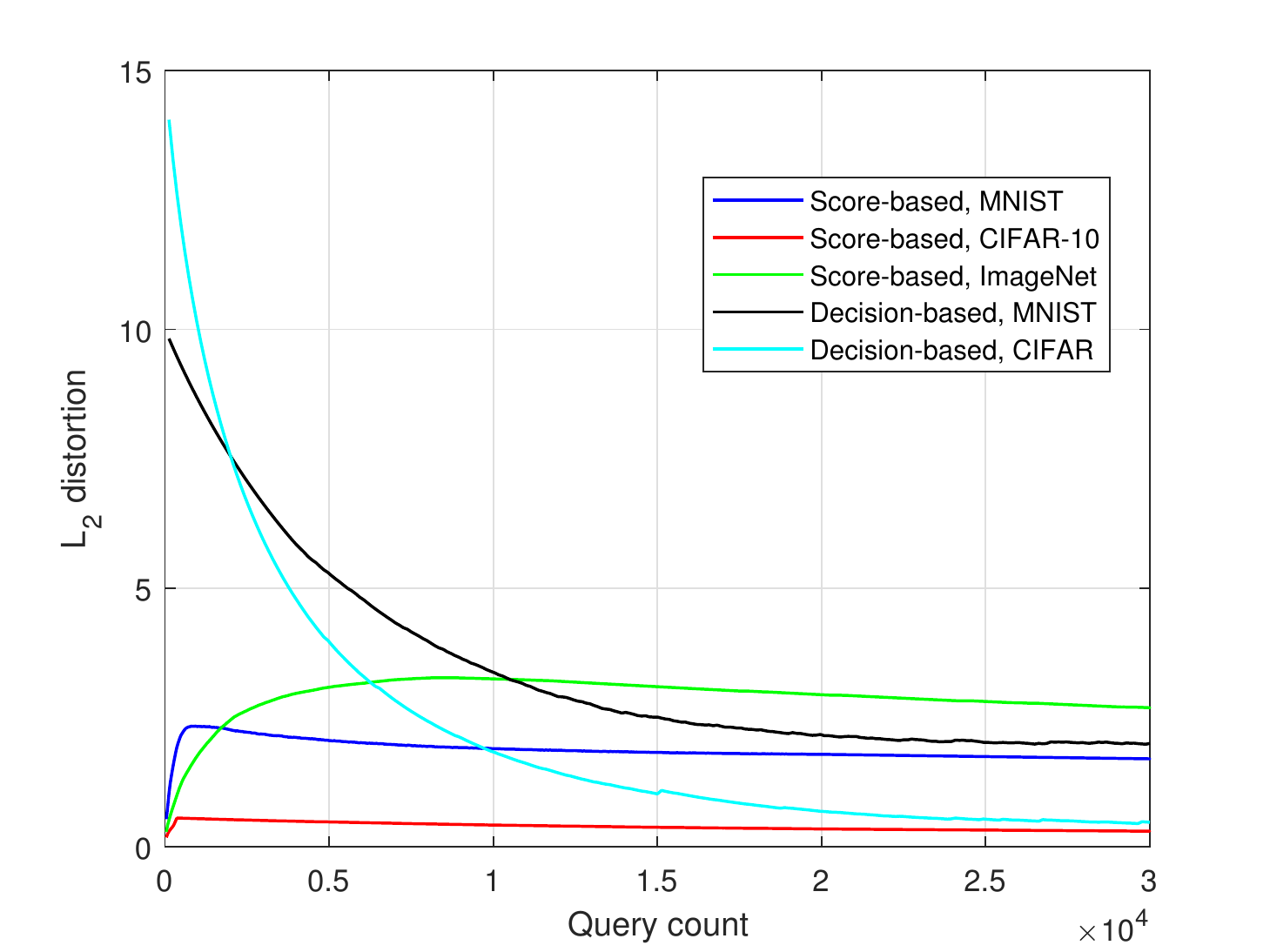}  
\end{tabular}}
\caption{Convergence of the ZO-ADMM attack.}
  \label{conv_score}
\end{figure}

In Fig. \ref{conv_score}, we demonstrate the convergence of the proposed ZO-ADMM targeted black-box attack, where  
the average $\ell_2$ distortion of 9 targeted adversarial examples versus the query number 
is presented. 
As we can see, since we initialize the adversarial distortion from zeros, the  score-based ZO-ADMM attack
increases   $\ell_2$ distortion  until a successful adversarial example is found. After that, it tries to decrease the  $\ell_2$ distortion but keeps the target label unchanged. 
For the decision-based attack, Fig \ref{conv_score} shows that the $\ell_2$ distortion is initially large  as  ZO-ADMM   starts from an image in the target class instead of the original image. The resulting $\ell_2$ distortion  then decreases as the query number increases.
We highlight that the ZO-ADMM attack is able to reach the successful attack with hundreds of queries on MNIST or CIFAR-10 and tens of thousands of queries on ImageNet, which significantly outperforms the ZOO attack. 
Besides Fig. \ref{conv_score} demonstrating the  $\ell_2$ distortion versus query number, we present the $\ell_2$ distortion versus ADMM iteration number in the supplementary material and similar results can be drawn.

\subsection{Evaluation for Various $\ell_p$ Norms}

In the previous experiments, 
we mainly consider the case of  $D(\boldsymbol{z}) = \| \boldsymbol{z} \|_2^2$ for a fair comparison with other white-box and black-box algorithms. However, we highlight that the ZO-ADMM method is able to optimize various $\ell_p$ norms, not only $\ell_2$ norm.
In Table \ref{table_three_norm}, we present the experimental results for different $\ell_p$ norms when solving problem \eqref{eq: z_step_sim}. Here we focus on 
 three score-based black-box attacks with ZO-ADMM by minimizing the $\ell_0$, $\ell_1$ and $\ell_2$ distortion, respectively. As we can see, our proposed method is well adapted to different $\ell_p$ norms in the design of black-box adversarial examples.

\begin{table} [t]
 \centering
  \caption{Performance evaluation of the ZO-ADMM attacks on MNIST for different $\ell_p$ norms.}
  \label{table_three_norm}
 \scalebox{1.0}[1.0]{
   \begin{threeparttable}
\begin{tabular}{c|c|c|c|c}
    \hline
\toprule[1pt]
 Attack method & ASR &  $\ell_0$ & $\ell_1$ &$\ell_2$  \\
\midrule[1pt]
ZO-ADMM $\ell_0$ & 100\% & \textbf{ 18.5} & 12.6 &  9.72 \\
ZO-ADMM $\ell_1$ &  100\% & 465 & \textbf{ 10.5} & 2.71  \\
ZO-ADMM $\ell_2$ & 100\% & 483 & 22.09 & \textbf{1.93}  \\
\bottomrule[1pt]
  \end{tabular}
\end{threeparttable}}  
\end{table}

\section{Conclusion}

In this paper, we propose a general framework to design \textcolor{black}{norm-ball bounded} black-box adversarial examples by leveraging  an operator splitting method (namely, ADMM), together with the gradient-free operations including random gradient estimation and Bayesian optimization. The proposed framework can be applied to both score-based and decision-based settings. Compared  to  state-of-the-art  black-box  attacks,  our  approach  achieves better query efficiency without losing   the attack performance in terms of  attack success rate as well as $\ell_p$-norm distortion.

\section*{Acknowledgement}
This work is partly supported by the National Science Foundation CNS-1932351.

{\small
\bibliographystyle{ieee_fullname}
\bibliography{egbib}
}
\clearpage
\appendix
\setcounter{table}{0}
\renewcommand{\thetable}{A\arabic{table}}
\setcounter{figure}{0}
\renewcommand{\thefigure}{A\arabic{figure}}
\begin{appendices}
\section{Solutions for different $\ell_p$ norms in $\bm z$-step}
In problem (11), we set $D(\boldsymbol{z}) = \| \boldsymbol{z} \|_2^2$ to measure the similarity between the legitimate image and the adversarial example. But $D(\boldsymbol{z})$  can also take other $\ell_p$ norms and the solutions in $z$-step can be obtained with minor modifications. 
In the following, we show the $z$-step solutions\footnote{We do not investigate the case of $\ell_\infty$ norm since the constraint $\|\bm z\|_\infty \leq \epsilon $ on the $\ell_\infty$ norm is already taken into consideration.} for $D(\boldsymbol{z}) = \| \boldsymbol{z} \|_0$, $D(\boldsymbol{z}) = \| \boldsymbol{z} \|_1$, and $D(\boldsymbol{z}) = \| \boldsymbol{z} \|_1 + \frac{\beta}{2} \| \boldsymbol{z} \|_2^2 $, derived from proximal operators 
which are     applicable and   well-suited to problems of substantial recent interest involving large or high-dimensional datasets.

\subsection{Solutions for $\ell_0$ norm}
If $D(\boldsymbol{z}) = \| \boldsymbol{z} \|_0$, the solution to problem (11) can be obtained as follows,
\begin{align}
\small
[\mathbf z^{k+1}]_i =    \left \{
    \begin{array}{ll}
       \min \{ 1- \left [\mathbf x_0 \right ]_i, \epsilon \}  &  
      \text{if} \ \ c_i  
        > \min \{ 1- \left [ \mathbf x_0  \right ]_i, \epsilon \}  \\
     \max \{ -  \left [ \mathbf x_0  \right ]_i, -\epsilon \}   &  
      \text{if} \ \  c_i 
     <  \max \{ -  \left [ \mathbf x_0  \right ]_i, -\epsilon \}   \\
      c_i  & \text{otherwise,}
    \end{array}
    \right.
\end{align}
where
\begin{align}
{c_i} = \left\{ {\begin{array}{*{20}{c}}
{{a_i}}&{{\rm{if}}\;a_i^2 > \frac{{2\gamma }}{\rho }}\\
0&{{\rm{otherwise}}}
\end{array}} \right.
\end{align}

\subsection{Solutions for $\ell_1$ norm}
If $D(\boldsymbol{z}) = \| \boldsymbol{z} \|_1$, the solution to problem (11) can be obtained as below,
\begin{align}
\small
[\mathbf z^{k+1}]_i =    \left \{
    \begin{array}{ll}
       \min \{ 1- \left [\mathbf x_0 \right ]_i, \epsilon \}  &  
      \text{if} \ \ (a_i-\frac{\gamma}{\rho})_+ - (- a_i - \frac{\gamma}{\rho})_+  \\
      &  > \min \{ 1- \left [ \mathbf x_0  \right ]_i, \epsilon \}  \\
     \max \{ -  \left [ \mathbf x_0  \right ]_i, -\epsilon \}   &  
      \text{if} \ \  (a_i-\frac{\gamma}{\rho})_+ - (- a_i - \frac{\gamma}{\rho})_+ \\
     &<  \max \{ -  \left [ \mathbf x_0  \right ]_i, -\epsilon \}   \\
       (a_i-\frac{\gamma}{\rho})_+ \\ - (- a_i - \frac{\gamma}{\rho})_+  & \text{otherwise,}
    \end{array}
    \right.
\end{align}
where $(x)_+ =x$ if $x \geq 0$ and 0 otherwise.

\subsection{Solutions for combination of $\ell_1$ and $\ell_2$ norm}
If $D(\boldsymbol{z}) = \| \boldsymbol{z} \|_1 +  \frac{\beta}{2} \| \boldsymbol{z} \|_2^2$, which is also know as elastic net regularization, the solution to problem (11) can be obtained through,
\begin{align}
\small
[\mathbf z^{k+1}]_i =    \left \{
    \begin{array}{ll}
       \min \{ 1- \left [\mathbf x_0 \right ]_i, \epsilon \}  &  
      \text{if} \ \ \frac{1}{1+\frac{\gamma \beta}{\rho}} ( (a_i-\frac{\gamma}{\rho})_+ \\ &- (- a_i - \frac{\gamma}{\rho})_+  )  \\
      & > \min \{ 1- \left [ \mathbf x_0  \right ]_i, \epsilon \}  \\
     \max \{ -  \left [ \mathbf x_0  \right ]_i, -\epsilon \}   &  
      \text{if} \ \  \frac{1}{1+\frac{\gamma \beta}{\rho}} ( (a_i-\frac{\gamma}{\rho})_+  \\ & - (- a_i - \frac{\gamma}{\rho})_+  ) \\
     &<  \max \{ -  \left [ \mathbf x_0  \right ]_i, -\epsilon \}   \\
      \frac{1}{1+\frac{\gamma \beta}{\rho}} ( (a_i-\frac{\gamma}{\rho})_+ \\ - (- a_i - \frac{\gamma}{\rho})_+  )  & \text{otherwise,}
    \end{array}
    \right.
\end{align}

\section{Derivation for maximizing EI}
EI can be transformed as follows,
\begin{align}\label{eq: gx_close}
    \mathrm{EI}(\boldsymbol{\delta})  & \overset{l^\prime = 
    \frac{l(\boldsymbol{\delta}) - \mu}{\sigma} }{=}  \mathbb E_{l^\prime}
    \left [
  ( l^+ - l^\prime \sigma - \mu ) \mathcal I \left (
  l^\prime \leq \frac{l^+ - \mu}{\sigma}
  \right )
    \right ] \nonumber \\
     & = (l^+ - \mu) \Phi \left ( \frac{l^+ - \mu}{\sigma} \right ) - \sigma E_{l^\prime}
    \left [ 
    l^\prime \mathcal I \left (
  l^\prime \leq \frac{l^+ - \mu}{\sigma}
  \right )
    \right ] \nonumber \\
    &  = (l^+ - \mu) \Phi \left ( \frac{l^+ - \mu}{\sigma} \right ) - \sigma \int_{-\infty}^{\frac{l^+ - \mu}{\sigma}} l^\prime \phi (l^\prime) d l^\prime \nonumber \\
    & =  (l^+ - \mu) \Phi \left ( \frac{l^+ - \mu}{\sigma} \right ) + \sigma\phi\left (
    \frac{l^+ - \mu}{\sigma}
    \right ),
\end{align}

\section{BO-ZO-ADMM}
In BO-ZO-ADMM, BO is used to obtain a query-efficient attack solution (at early ADMM iterations) for initializing the ZO method, which can further minimize  the  adversarial distortion (at later ADMM iterations). Additional experiments showed that when reaching the same $\ell_2$ distortion as  ZO-ADMM,   BO-ZO-ADMM    requires $380$   queries on   MNIST and $320$ queries on CIFAR-10, outperforming {493 and 421 queries in Table\,1.}

\section{Comparison with AutoZoom and Boundary method}

For the comparison with AutoZoom \cite{tu2018autozoom}, we report the averaged number of queries for attacking 500images at the same $\ell_2$ distortion level for MNIST, CIFAR-10,and ImageNet in Table \ref{tab: autozoom}.  As we can see, the proposed ZO-ADMM method is more query-efficient, while it is worth noting that AutoZOOM produces adversarial  perturbation  in  low-dimensional  latent  space,and thus saves more computation cost.

\begin{table}[htb]
\centering
\scalebox{0.73}[0.73]{
\begin{threeparttable}
\caption{Comparison to AutoZOOM in attack success rate (ASR) and query \#.}
\label{tab: autozoom}
{\small
\begin{tabular}{c|c|c|c|c|c|c}
\hline%
 & \multicolumn{2}{c|}{ MNIST} &\multicolumn{2}{c|}{  CIFAR-10} & \multicolumn{2}{c}{ ImageNet} \\
\hline
& ASR &  \# of Query & ASR & \# of Query  &ASR & \# of Query \\
\hline
AutoZoom & 100\% & 1821 &  99.2\% &  1639  & 98.3\% & 43547 \\
ZO-ADMM &  100\% &  562 &  99\% & 492 & 99\% &16390 \\
\hline
\end{tabular}}
\end{threeparttable}}
\end{table}

In Table \ref{tab: boundary}, we show the comparison of ZO-ADMM method with the Query-limited \cite{ilyas2018blackbox} and Boundary methods \cite{brendel2017decision} in terms of query number and $\ell_p$ norms on ImageNet. 

\begin{table}[htb]
\centering
\scalebox{0.7}[0.7]{
\begin{threeparttable}
\caption{Experimental results on ImageNet}
\label{tab: boundary}
\begin{tabular}{c|c|c|c|c|c|c}
\hline%
Settings & Methods & ASR & $\ell_1$ & $\ell_2$ & $\ell_\infty$ & Query \# \\
\hline
\multirow{2}{*}{ \shortstack{Score-\\based}}  & Query-limited [18] & 98\% & 1251 & 4.8 & 0.049 & $3.4 \times 10^{5} $\\
&  ZO-ADMM  & 97\% & 785 & 3.5 & 0.039 &  $1.6 \times 10^{5} $ \\
\hline
\multirow{2}{*}{ \shortstack{Decision-\\based}} & Boundary [20] & 85\% & 1120 & 3.99 & 0.045 &   $2.2 \times 10^{6} $   \\
&  ZO-ADMM  & 93\%  & 962 & 3.92  & 0.042  & $1.5 \times 10^{6} $ \\
\hline
\end{tabular}
\end{threeparttable}}
\end{table}

\section{Convergence of the ZO-ADMM attack}

\begin{figure}[h]
\centerline{
\begin{tabular}{c}
\includegraphics[width=.45\textwidth]{figs/conv_score_decision-eps-converted-to.pdf}   \\
  \footnotesize{(a) $\ell_2$ norm v.s. query number} \\
\includegraphics[width=.45\textwidth]{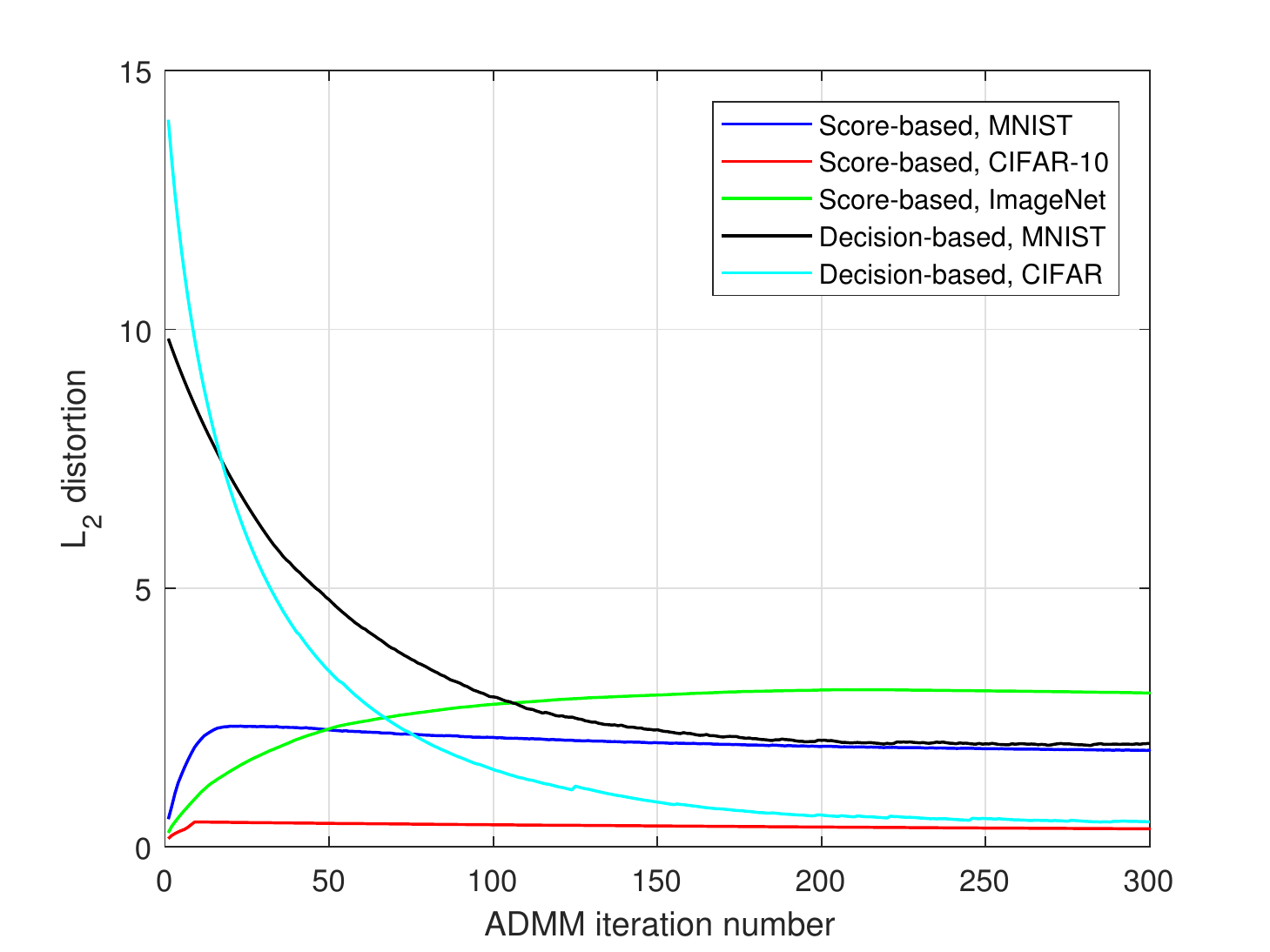}  
\\
 \footnotesize{(b) $\ell_2$ norm v.s. ZO-ADMM iteration number]} 
\end{tabular}}
\caption{Convergence of the ZO-ADMM attack.}
  \label{fig: conv_score}
\end{figure}

\begin{figure}
\centering
  \includegraphics[width=1.0\linewidth]{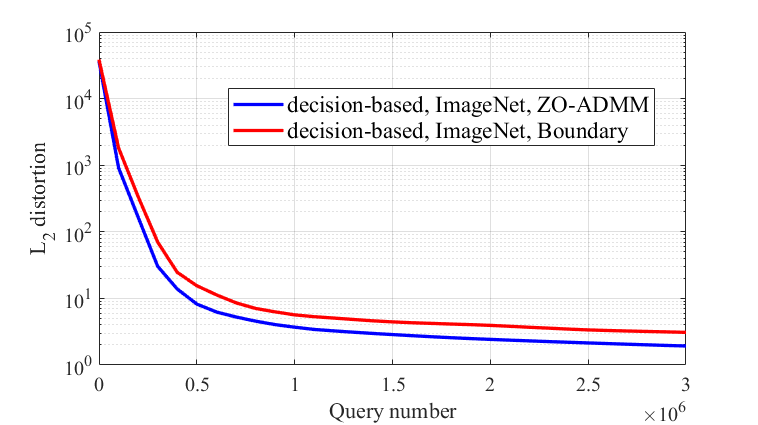}
  \caption{\footnotesize{$\ell_2$ distortion of decision-based attack vs queries on ImageNet.}} 
  \label{fig: convergence_imagenet}
  \vspace{-0.2in}
\end{figure}

Figure \ref{fig: conv_score} shows the convergence of the ZO-ADMM attack v.s. query number or ADMM iteration number. Figure \ref{fig: convergence_imagenet} shows the convergence comparison of the ZO-ADMM method and the Boundary method. 
 
\section{Examples for the decision-based  ZO-ADMM attack}

In the following, we provide more adversarial examples generated by the proposed ZO-ADMM decision-based black-box attack.

\begin{figure}[hbt]
\centering
	\includegraphics[ scale=0.6]{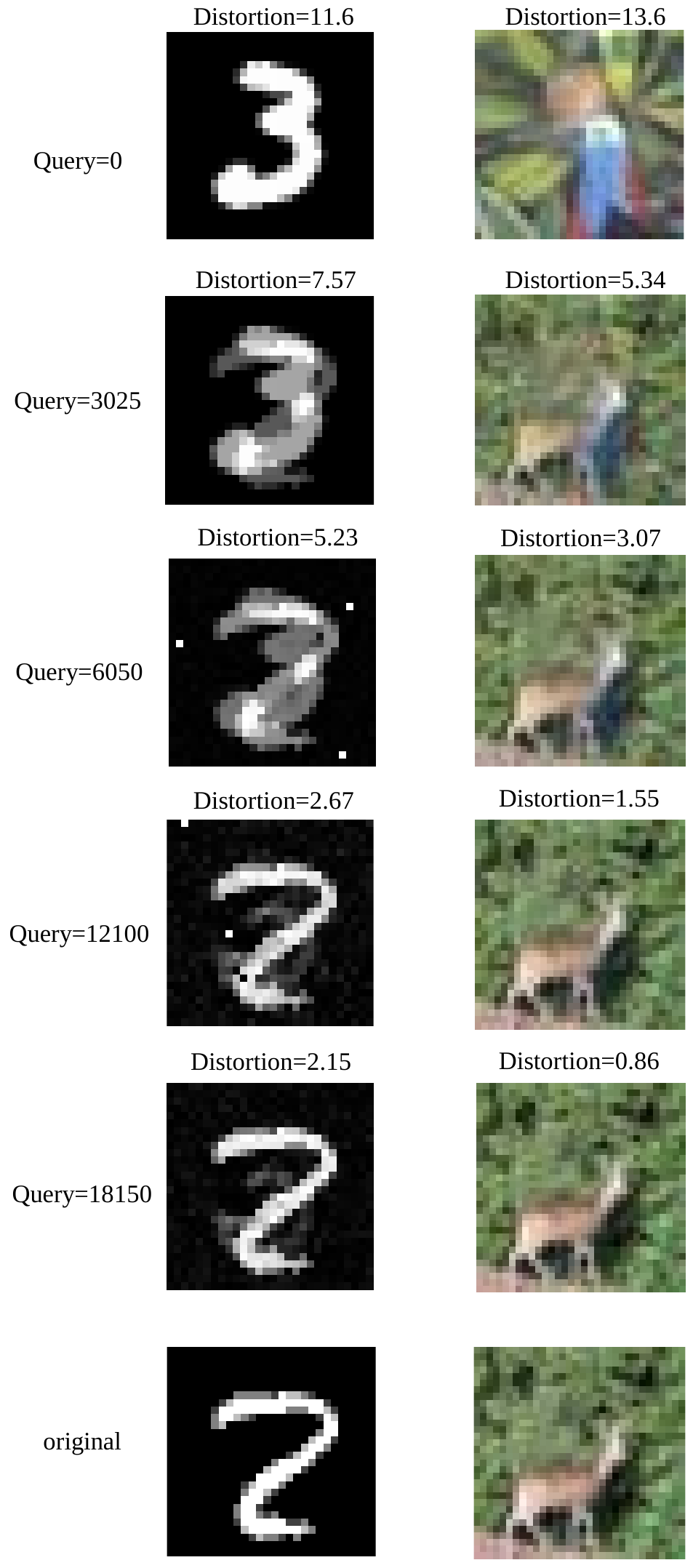}
 \caption{\textcolor{black}{Adversarial examples generated by the proposed decision-based black-box attack with ZO-ADMM} on MNIST and CIFAR-10. } \label{envo_mnist_cifar_2}\vspace{30pt}
\end{figure}

\begin{figure}[htb]
	\center
	\includegraphics[ scale=0.4]{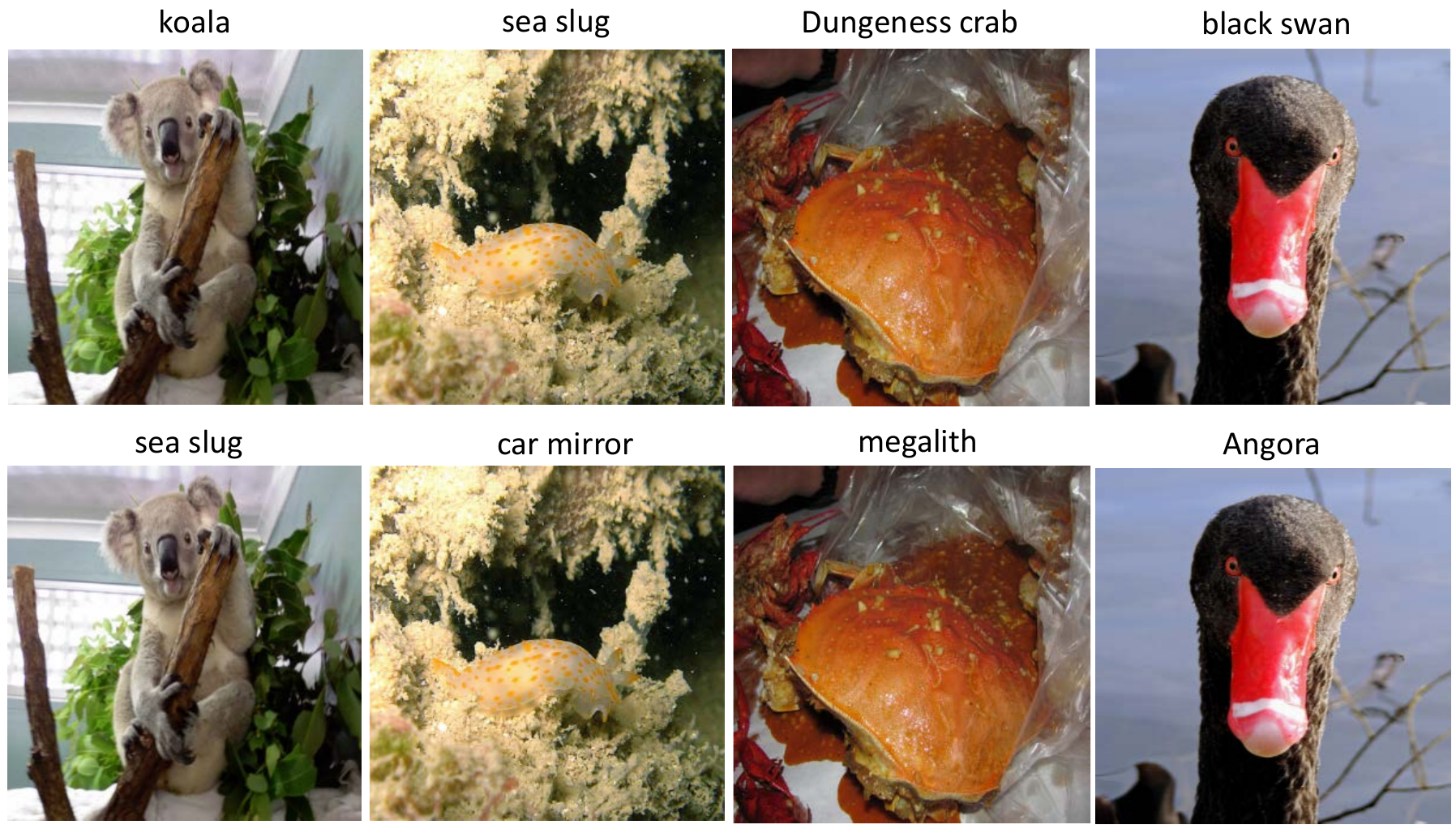}
	\caption{Adversarial examples on ImageNet. The original images are on the top row and their corresponding adversarial examples are shown on the bottom row with target labels. }
	\label{examples_imagenet_2}
\end{figure}

\end{appendices}
\end{document}